%% file: main.tex
\documentclass[conference]{IEEEtran}
\usepackage{times}

\usepackage[numbers]{natbib}
\usepackage{multicol}
\usepackage[bookmarks=True]{hyperref}

\hypersetup{
	colorlinks=true,
	linkcolor=orange,
	filecolor=magenta,      
	urlcolor=orange,
	citecolor=orange,
}
\usepackage[utf8]{inputenc} %
\usepackage[T1]{fontenc}    %
\usepackage{amsfonts}       %
\usepackage{nicefrac}       %
\usepackage{microtype}      %

\usepackage{times}
\usepackage{graphicx}
\usepackage{caption}
\usepackage{listings}[language=Python]
\usepackage{color}
\usepackage[table,xcdraw,dvipsnames,usenames]{xcolor}
\usepackage{booktabs}
\usepackage{multirow}
\usepackage{natbib}
\usepackage[capitalise, nameinlink]{cleveref}
\definecolor{codegreen}{rgb}{0,0.6,0}
\definecolor{codegray}{rgb}{0.5,0.5,0.5}
\definecolor{codepurple}{rgb}{0.58,0,0.82}
\definecolor{backcolour}{rgb}{0.95,0.95,0.92}

\lstdefinestyle{mystyle}{
    backgroundcolor=\color{backcolour},
    commentstyle=\color{codegreen},
    keywordstyle=\color{magenta},
    numberstyle=\tiny\color{codegray},
    stringstyle=\color{codepurple},
    basicstyle=\ttfamily\footnotesize,
    breakatwhitespace=false,         
    breaklines=true,                 
    captionpos=b,                    
    keepspaces=true,                 
    numbers=left,                    
    numbersep=5pt,                  
    showspaces=false,                
    showstringspaces=false,
    showtabs=false,                  
    tabsize=2
}
\definecolor{dgreen}{rgb}{0,.8,0.8}
\definecolor{dyellow}{rgb}{.7,.7,0}
\definecolor{dred}{rgb}{1,0,0}
\definecolor{dblue}{rgb}{0,0,0.7}
\definecolor{dorange}{rgb}{0.9,0.5,0.1}
\definecolor{dgray}{rgb}{0.5,0.5,0.5}

\usepackage{amssymb}
\usepackage{pifont}%
\usepackage{colortbl}
\usepackage{wrapfig}
\usepackage{soul}
\usepackage{wrapfig}
\newcommand*\colourcross[1]{%
  \expandafter\newcommand\csname #1xmark\endcsname{\textcolor{#1}{\ding{55}}}%
}
\newcommand*\colourcheck[1]{%
  \expandafter\newcommand\csname #1check\endcsname{\textcolor{#1}{\ding{52}}}%
}
\colourcross{red}
\colourcheck{green}

\newcounter{savefootnote}
\newcounter{symfootnote}
\newcommand{\symfootnote}[1]{%
   \setcounter{savefootnote}{\value{footnote}}%
   \setcounter{footnote}{\value{symfootnote}}%
   \ifnum\value{footnote}>8\setcounter{footnote}{0}\fi%
   \let\oldthefootnote=\thefootnote%
   \renewcommand{\thefootnote}{\fnsymbol{footnote}}%
   \footnote{#1}%
   \let\thefootnote=\oldthefootnote%
   \setcounter{symfootnote}{\value{footnote}}%
   \setcounter{footnote}{\value{savefootnote}}%
}

\newcommand{\ie}{i.e., }
\newcommand{\eg}{e.g., }

\usepackage{calc}
\newlength\myheight
\newlength\mydepth
\settototalheight\myheight{Xygp}
\newcommand{\methodfancy}{Octo} %
\newcommand{\methodfancyreverse}{Octo} %
\newcommand{\method}{Octo}
\newcommand{\ndatasets}{25}
\newcommand{\nsetups}{9}
\newcommand{\ninstitutions}{4}
\newcommand{\ntrajs}{800k}
\newcommand{\tinynparams}{10M}
\newcommand{\smallnparams}{27M}
\newcommand{\largenparams}{93M}

\newcommand{\website}{\url{https://octo-models.github.io}}

\begin{document}

\title{\methodfancyreverse: An Open-Source Generalist Robot Policy\vspace{-0.5cm}}

\author{
\hspace{-1cm}\textbf{\method{} Model Team}\\[0.2cm] 
\hspace{-1cm}Dibya Ghosh$^{\ast,1}$ \hspace{0.2cm} Homer Walke$^{\ast,1}$ \hspace{0.2cm} Karl Pertsch$^{\ast,1,2}$ \hspace{0.2cm} Kevin Black$^{\ast,1}$ \hspace{0.2cm} Oier Mees$^{\ast,1}$\\[0.2cm]
\hspace{-0.5cm}Sudeep Dasari$^{3}$ \hspace{0.15cm} Joey Hejna$^{2}$ \hspace{0.15cm} Tobias Kreiman$^{1}$ Ria Doshi$^{1}$\hspace{0.15cm} Charles Xu$^{1}$ \hspace{0.15cm} Jianlan Luo$^{1}$ \hspace{0.15cm} You Liang Tan$^{1}$\\
\hspace{-0.7cm}Lawrence Yunliang Chen$^{1}$\hspace{0.15cm} Pannag Sanketi$^{4}$\hspace{0.15cm} Quan Vuong$^{4}$\hspace{0.15cm} Ted Xiao$^{4}$ \hspace{0.15cm} Dorsa Sadigh$^{2}$ \hspace{0.15cm} Chelsea Finn$^{2}$ \hspace{0.15cm} Sergey Levine$^{1}$\\[0.2cm]
\hspace{-1cm}$^{1}$UC Berkeley $^{2}$Stanford $^{3}$Carnegie Mellon University ${}^{4}$Google Deepmind\\
\website
}

\makeatletter
\let\@oldmaketitle\@maketitle%
\renewcommand{\@maketitle}{\@oldmaketitle%
  \begin{center}
  \captionsetup{type=figure}
  \includegraphics[width=\textwidth]{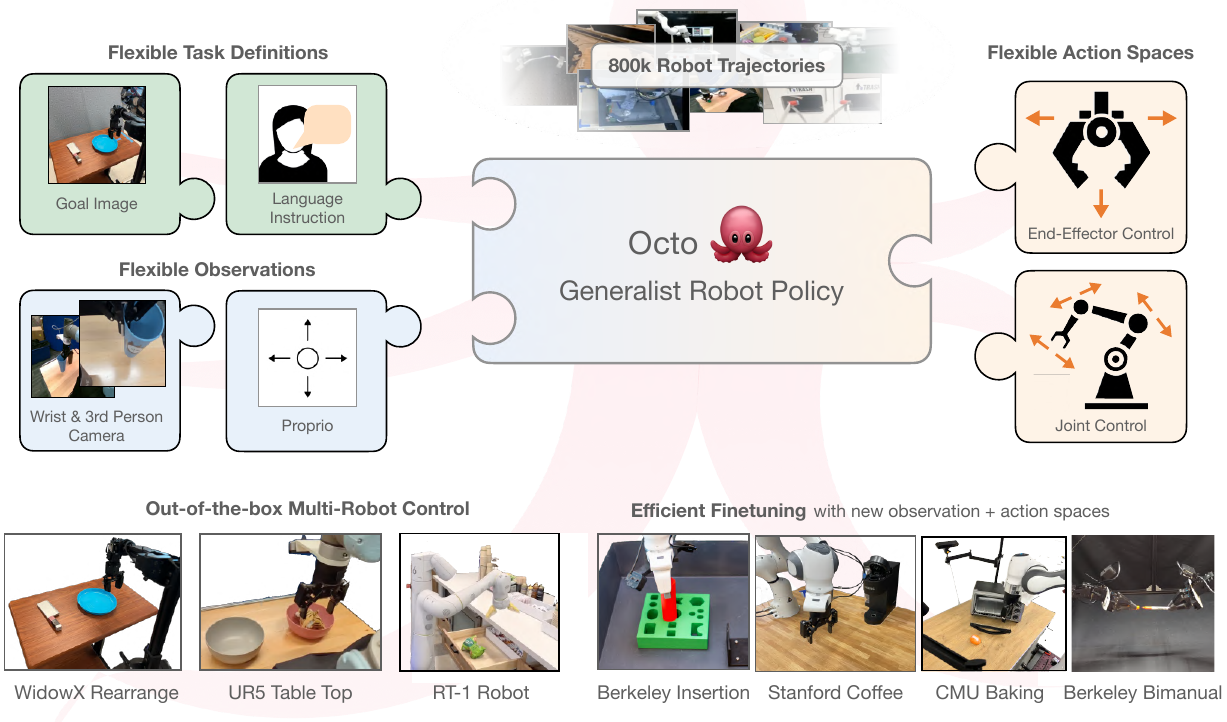}
    \captionof{figure}{\small We introduce \method{}, %
    an open-source, generalist policy for robotic manipulation. \method{} is a transformer-based policy pretrained on \ntrajs{} diverse robot episodes from the Open X-Embodiment dataset~\citep{open_x_embodiment_rt_x_2023}. It supports flexible task and observation definitions and can be quickly finetuned to new observation and action spaces.} 
    \vspace{-1.5em}
    \label{fig:teaser}
    \end{center}
}
\makeatother

\maketitle
\addtocounter{figure}{-1}

\input{sections/01_abstract}

\IEEEpeerreviewmaketitle

\input{sections/02_intro}

\input{sections/02.5_related_work}
\input{sections/03_approach}
\input{sections/04_experiments}

\input{sections/05_conclusion}

\bibliographystyle{plainnat}
\bibliography{references}

\clearpage
\input{sections/06_appendix}

\end{document}

%% file: sections/01_abstract.tex
\begin{abstract}

Large policies pretrained on diverse robot datasets have the potential to transform robotic learning: instead of training new policies from scratch, such generalist robot policies may be finetuned with only a little in-domain data, yet generalize broadly.
However, to be widely applicable across a range of robotic learning scenarios, environments, and tasks, such policies need to handle diverse sensors and action spaces, accommodate a variety of commonly used robotic platforms, and finetune readily and efficiently to new domains.
In this work, we aim to lay the groundwork for developing open-source, widely applicable, generalist policies for robotic manipulation. As a first step, we introduce \methodfancy, a large transformer-based policy trained on \ntrajs{} trajectories from the Open X-Embodiment dataset, the largest robot manipulation dataset to date.
It can be instructed via language commands or goal images and can be effectively finetuned to robot setups with new sensory inputs and action spaces within a few hours on standard consumer GPUs. 
In experiments across \nsetups{} robotic platforms, we demonstrate that \method{} serves as a versatile policy initialization that can be effectively finetuned to new observation and action spaces. We also perform detailed ablations of design decisions for the \method{} model, from architecture to training data, to guide future research on building generalist robot models.

\end{abstract}

%% file: sections/02_intro.tex
\section{Introduction}
\label{sec:intro}
\let\thefootnote\relax\footnotetext{$^*$Lead authors, ordered alphabetically, see \cref{sec:contributions} for list of contributions. \\Correspondence to \href{mailto:dibya.ghosh@berkeley.edu,homer_walke@berkeley.edu,pertsch@berkeley.edu,kvablack@berkeley.edu,oier.mees@eecs.berkeley.edu}{\scriptsize\texttt{\{dibya.ghosh, homer\_walke, pertsch, kvablack, oier.mees\}@berkeley.edu}}}
\setcounter{footnote}{0}
The common approach for robotic learning is to train policies on datasets collected for the specific robot and task at hand. Learning from scratch in this way requires significant data collection effort for each task, and the resulting policies usually exhibit only narrow generalization. In principle, collected experience from other robots and tasks offers a possible solution, exposing models to a diverse set of robotic control problems that may improve generalization and performance on downstream tasks.
However, even as general-purpose models become ubiquitous in natural language \citep{openaiGPT4TechnicalReport2023,touvronLLaMAOpenEfficient2023}) and computer vision \citep{rombachHighResolutionImageSynthesis2022, kirillovSegmentAnything2023}, it has proven challenging to build the analogous ``general-purpose robot model'' that can control many robots for many tasks. Training a unified control policy in robotics presents unique challenges, requiring handling different robot embodiments, sensor setups, action spaces, task specifications, environments, and compute budgets.

Towards this direction, several works have proposed robotic foundation models that directly map robot observations to actions and provide zero-shot or few-shot generalization to new domains and robots.
We broadly refer to these models as ``generalist robot policies'' (GRPs), 
emphasizing their ability to perform low-level visuomotor control across tasks, environments, and robotic systems~\citep{reed2022generalist,bousmalis2023robocat,driess2023palm,zitkovich2023rt,brohan2022rt,shah2023vint,scaleblog,wayveblog,hu2023gaia1,yang2023polybot,kumar2023robohive}. For example, the GNM model~\citep{shahGNMGeneralNavigation2023} generalizes across different robotic navigation scenarios, the RoboCat model~\citep{bousmalis2023robocat} handles different robot embodiments for goal-conditioned tasks, and the RT-X model~\citep{open_x_embodiment_rt_x_2023} performs language-conditioned manipulation across five robot embodiments. 
Although these models represent significant steps toward a true ``general-purpose robot model,'' they have been limited in multiple important aspects: they typically constrain downstream users to a pre-defined and often restrictive set of input observations, \eg a single camera stream; they lack support for effective finetuning to new domains; and importantly, the largest of these models are not available to the general public. 

We design a system for pretraining generalist robot policies more suitable for the diversity of interfaces in downstream robotic applications. The core of our model is a transformer architecture that maps arbitrary input tokens (created from observations and tasks) to output tokens (then decoded into actions), which can be trained on a diverse dataset of robots and tasks. With no additional training, this policy can accept different camera configurations (e.g., workspace or wrist cameras), can control different robots, and can be guided via either language commands or goal images --- all by simply changing which tokens are fed into the model. Most importantly, the model can be adapted to new robot setups with new sensory inputs, action spaces, or morphologies by adding appropriate adapters and finetuning with a small target domain dataset and an accessible compute budget.

Our primary contribution is \methodfancy{}, a transformer-based policy pretrained on the largest robot manipulation dataset to date: \ntrajs{} robot demonstrations from the Open X-Embodiment dataset~\citep{open_x_embodiment_rt_x_2023}. \method{} is the first GRP that can be effectively finetuned to new observations and action spaces and the first generalist robot manipulation policy that is fully open-source, including the training pipeline, model checkpoints, and data. Finally, while the individual components that comprise \method{} --- a transformer backbone, support for both language and goal image specification, and a diffusion head to model expressive action distributions --- have been discussed in prior work, the particular combination of these components into a powerful generalist robot policy is unique and novel.

We demonstrate through extensive experiments on \nsetups{}~robots across \ninstitutions{}~institutions that our combined system leads to state-of-the-art performance for out-of-the-box multi-robot control for single and dual-arm manipulation tasks and that \method{} can be used as an effective initialization for finetuning to unseen setups with new observation and action spaces. In the process, we carefully study the effect of different design decisions when pretraining GRPs; we evaluate how the choice of data distribution, model architecture, and policy formulation affects the quality of the pretrained GRP. Our evaluation highlights the utility of scale and flexibility: our best models are those trained on the widest data mixtures, with the least restrictive inductive biases, and with policy objectives that can fit the diversity of behaviors in the pretraining data.

Along with this paper, we release all resources required to train, use, reproduce, and finetune an \method{} model. We provide pretrained \method{} model checkpoints with \smallnparams{} and \largenparams{} parameters that, out of the box, support multiple RGB camera inputs as well as both language and goal image task specification. We also provide scripts for finetuning these models on new domains, as well as our complete pretraining pipeline, including optimized data loaders, transformer implementations for multimodal inputs, and tools to monitor training progress.

%% file: sections/02.5_related_work.tex
\section{Related Work}
\label{sec:related_work}

\begin{figure*}[t]
    \centering
    \includegraphics[width=\textwidth]{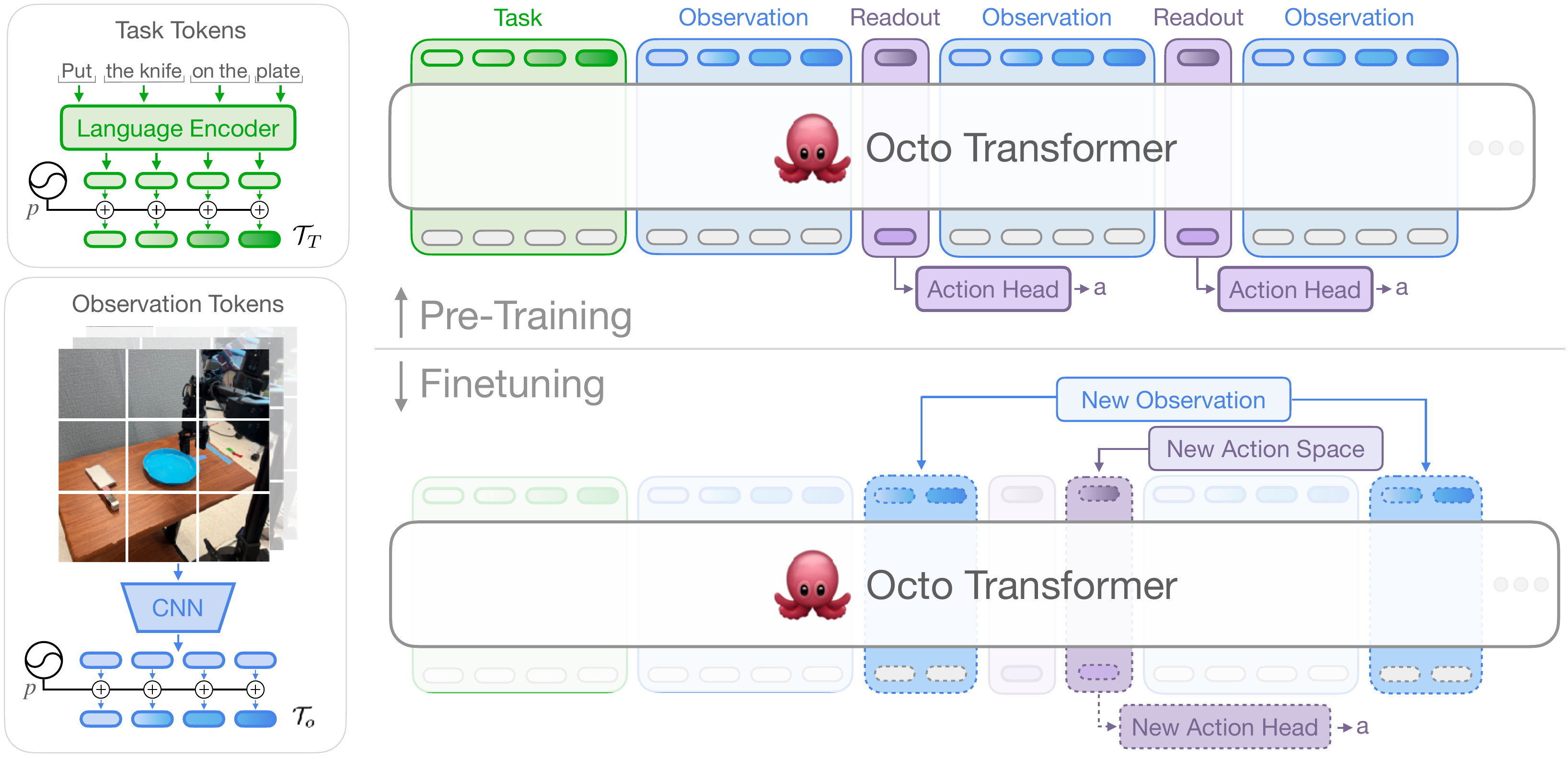}
    \caption{\textbf{Model architecture.} \textbf{Left}: \method{} tokenizes task descriptions (green) and input observations (blue) using a pretrained language model and a lightweight CNN, respectively. \textbf{Top}: The transformer backbone processes the sequence of task and observation tokens and produces readout tokens (purple) that get passed to output heads to produce actions. \textbf{Bottom}: The block-wise attention structure of the transformer allows us to add or remove inputs and outputs during finetuning: for example, we can add new observations (blue, dashed) or action spaces (purple, dashed) without modifying any pretrained parameters.}
    \label{fig:architecture}
\end{figure*}

Many works train policies using a large dataset of trajectories collected from a robot, from early efforts using autonomous data collection for scaling policy training~\citep{pinto2016supersizing,levine2018learning,kalashnikov2018qt,dasari2020robonet,finn2017deep,gupta2018robot} to more recent efforts that explore the combination of modern transformer-based policies with large demonstration datasets~\citep{brohan2022rt,pmlr-v202-jiang23b,zhao2023learning,fu2024mobile,shridhar2023perceiver,stone2023open}. These works primarily focus on a single embodiment, while \method{} trains policies on robot datasets assembled across \emph{multiple} embodiments, increasing the effective size of the training dataset and allowing finetuning to a range of robot setups. 

More recently, papers have focused on broadening the generalization abilities of robot policies. Multiple works leverage diverse non-robot data or pretrained vision-language foundation models to boost policy generalization to new scenes and tasks~\citep{stone2023open,zitkovich2023rt,yu2023scaling,chen2023genaug,huang2023voxposer,ahn2022can,singh2023progprompt,huang2023visual,bahl2023affordances,huang23avlmaps,black2023zero,bahl2022human,kwon2023language,chen2024vision,driess2023palm}. More closely related to \method{} are recent works that train robot policies across data from multiple robot embodiments: the GNM model~\citep{shah2023vint,shahGNMGeneralNavigation2023}
generalizes across robot navigation setups while RoboCat~\citep{bousmalis2023robocat} and RT-X~\cite{open_x_embodiment_rt_x_2023} control multiple single-arm manipulation robots. While these models deliver impressive policy learning results, a key issue is their lack of flexibility: they typically require users to stick to the sensory inputs and action space used during pretraining and do not support adaptation to new observation and action spaces. Furthermore, the largest models are not publicly accessible. \method{} differs from these works in multiple aspects: it is trained on a larger and more diverse robot data mix, it supports a wider range of downstream applications via efficient finetuning to new robot setups, and it is fully open source and reproducible.

\method{}'s design is inspired by several recent advances in robot imitation learning and scalable transformer training, including the use of denoising diffusion objectives~\citep{ho2020denoising} for action decoding~\citep{chi2023diffusionpolicy,ha2023scaling,sridhar2023nomad}, the prediction of ``action chunks'', \ie sequences of future actions~\citep{zhao2023learning,chi2023diffusionpolicy,fu2024mobile}, and model layouts and learning rate schedules inspired by the literature on scalable vision transformer training~\citep{dosovitskiy2020image,zhai2022scaling}. Our work is the first to leverage these approaches in the context of learning cross-embodied generalist policies and we find that they can lead to substantial performance improvements. In our evaluation, we present ablations to assess the importance of these components, alongside a more comprehensive list of what we found to be (un)important in Appendix \ref{sec:app:things_that_did_not_work}; we hope our findings are useful for future research on generalist policy learning.

A key ingredient for training generalist robot policies is robot training data. In contrast to vision and language data that can be scraped from the web, obtaining robot data at scale is challenging and often involves significant investments in hardware and human labor. There are multiple large robot navigation and autonomous driving datasets~\citep{geiger2012we,yu2020bdd100k,caesar2020nuscenes,sun2020scalability,shahGNMGeneralNavigation2023,karnan2022socially,triest2022tartandrive}. In recent years, there have also been multiple efforts for building robot \emph{manipulation} datasets of increasing scale and diversity, either collected via scripted and autonomous policies~\citep{dasari2020robonet, kalashnikov2018qt,kalashnikov2021scaling,cabi2019,pinto2016supersizing,gupta2018robot} or human teleoperation~\citep{mandlekar2018roboturk,mandlekar2023mimicgen,ebert2021bridge,walke2023bridgedata,jang2022bc,brohan2022rt,fang2023rh20t,bharadhwaj2023roboagent,rosetebeas2022latent,mees2023grounding,shafiullah2023dobbe}. \method{} is trained on the Open X-Embodiment dataset~\citep{open_x_embodiment_rt_x_2023}, a recent effort that pooled many of these aforementioned robot datasets. The Open-X dataset contains approximately 1.5M robot episodes, of which we curate \ntrajs{} for \method{} training. We note that the RT-X model \citep{open_x_embodiment_rt_x_2023} used a more restricted subset of 350K episodes, so to the best of our knowledge, Octo is trained on the largest robotics manipulation demonstration dataset to date.

%% file: sections/03_approach.tex
\section{The Octo Model}
\label{sec:approach}
In this section, we describe the \method{} model, our open-source generalist robot policy that can be adapted to new robots and tasks --- including new sensory inputs and action spaces --- via finetuning. We discuss the key design decisions, training objectives, training dataset, and infrastructure.
The design of the \method{} model emphasizes flexibility and scale: it supports a variety of commonly used robots, sensor configurations, and actions while providing a generic and scalable recipe that can be trained on large amounts of data.
It also supports natural language instructions, goal images, observation histories, and multi-modal, chunked action prediction via diffusion decoding~\citep{chi2023diffusionpolicy}. Furthermore, we designed \method{} specifically to enable efficient finetuning to new robot setups, including robots with different action spaces and different combinations of cameras and proprioceptive information. This design was selected to make \method{} a flexible and broadly applicable generalist robot policy that can be utilized for a variety of downstream robotics applications and research projects.

\subsection{Architecture}
\label{sec:arch}

At its core, \method{} is a transformer-based policy $\pi$. It consists of three key parts: \textbf{input tokenizers} that transform language instructions $\ell$, goals $g$, and observation sequences $o_1, \dots, o_H$ into tokens $\big[\mathcal{T}_l, \mathcal{T}_g, \mathcal{T}_o\big]$ (\cref{fig:architecture}, left); a \textbf{transformer backbone} that processes the tokens and produces embeddings $e_l, e_g, e_o = T(\mathcal{T}_l, \mathcal{T}_g, \mathcal{T}_o)$ (\cref{fig:architecture}, top); and \textbf{readout heads} $R(e)$ that produce the desired outputs, \ie actions $a$.

\paragraph{Task and observation tokenizers} We convert task definitions (\eg language instructions $\ell$ and goal images $g$) and observations $o$ (\eg wrist and third-person camera streams) into a common ``tokenized'' format using modality-specific tokenizers (see~\cref{fig:architecture}, left): 
\begin{itemize}
    \item \textbf{Language inputs} are tokenized, then passed through a pretrained transformer that produces a sequence of language embedding tokens. We use the \texttt{t5-base} (111M) model \citep{2020t5}.
    \item \textbf{Image observations and goals} are passed through a shallow convolution stack, then split into a sequence of flattened patches~\citep{dosovitskiy2020image}.
\end{itemize}

We assemble the input sequence of the transformer by adding learnable position embeddings $p$ to task and observation tokens and then arranging them sequentially $\big[\mathcal{T}_T, \mathcal{T}_{o,1}, \mathcal{T}_{o,2}, \dots\big]$.

\paragraph{Transformer backbone and readout heads} Once the inputs have been cast to a unified token sequence, they are processed by a transformer (see~\cref{fig:architecture}, top). This is similar to prior works that train transformer-based policies on sequences of observations and actions~\citep{wu2023masked,radosavovic2023robot}. The attention pattern of the \method{} transformer is block-wise masked: observation tokens can only attend causally to tokens from the same or earlier time steps $\mathcal{T}_{o, 0:t}$ as well as task tokens $\mathcal{T}_T$ (green). Tokens corresponding to non-existing observations are fully masked out (\eg a dataset without language instructions). This modular design enables us to add and remove observations or tasks during finetuning (see below). In addition to these input token blocks, we insert learned \emph{readout tokens} $\mathcal{T}_{R,t}$ (purple). A readout token at $\mathcal{T}_{R,t}$ attends to observation and task tokens before it in the sequence, but is not attended to by \textit{any} observation or task token --- hence, they can only passively read and process internal embeddings without influencing them. Readout tokens act similarly to the \texttt{[CLS]} token in BERT, serving as a compact vector embedding of the observation sequence thus far. A lightweight ``action head'' that implements the diffusion process is applied to the embeddings for the readout tokens. This action head predicts a ``chunk" of several consecutive actions, similar to prior work \cite{zhao2023learning, chi2023diffusionpolicy}. 

Our design allows us to flexibly add new task and observation inputs or action output heads to the model during downstream finetuning. When adding new tasks, observations, or loss functions downstream, we can wholly retain the pretrained weights for the transformer, only adding new positional embeddings, a new lightweight encoder, or the parameters of the new head as necessitated by the change in specification (see~\cref{fig:architecture}, bottom). This is in contrast to prior architectures \citep{brohan2022rt, shah2023vint}, where adding or removing an image input or changing the task specification would require re-initializing or re-training large components of the pre-trained model.

This flexibility is crucial to make \method{} a truly ``generalist'' model: since we cannot cover all possible robot sensor and action configurations during pretraining, being able to adapt \method's inputs and outputs during finetuning makes it a versatile tool for the robotics community. Prior model designs that use standard transformer backbones or fuse visual encoders with MLP output heads lock in the type and order of inputs expected by the model. In contrast, switching the observation or task for \method{} \emph{does not} require re-initializing most of the model.

\subsection{Training data}
\label{sec:train_data}

\begin{figure}[t]
\vspace{-0.3cm}
   \centering
   \includegraphics[width=\linewidth]{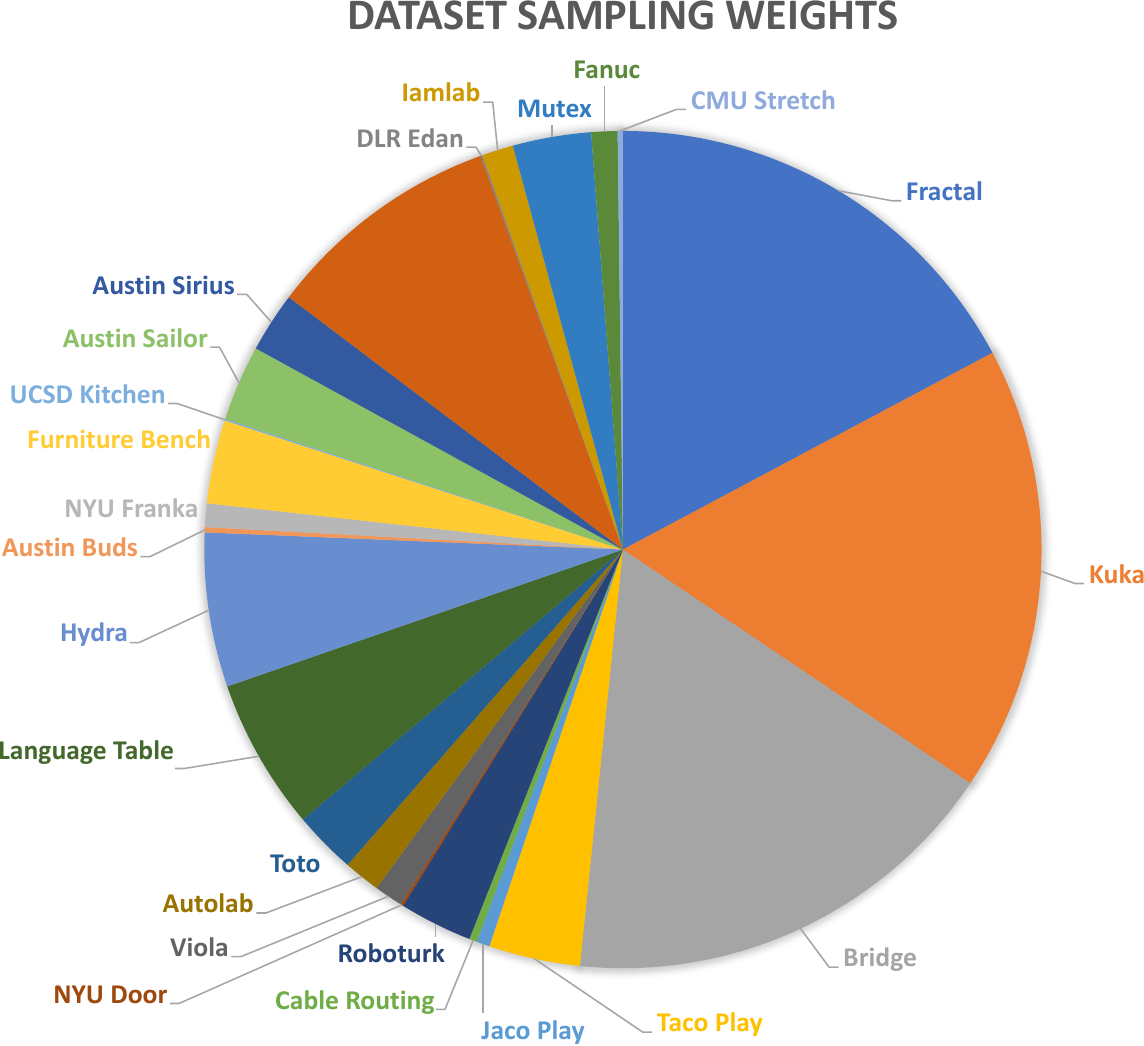}
   \caption{\textbf{Training dataset composition.}
    We curate a subset of \ndatasets{}~datasets from the Open X-Embodiment dataset that have image observations, end-effector actions, and show diverse behaviors. The pie chart visualizes the fractions that each dataset contributes to every training batch on average. The dataset weights are determined by the number of samples in each dataset with small modifications to balance dataset size and diversity (see \cref{sec:train_data} for details).}
    \label{fig:datamix}
\end{figure}
We train \method{} on a mixture of \ndatasets{} datasets from the Open X-Embodiment Dataset~\citep{open_x_embodiment_rt_x_2023}, a diverse collection of robot learning datasets. Our training mixture includes demonstration data of a variety of tasks from several robot embodiments and scenes. These datasets are heterogeneous not just in terms of the robot type, but also in the sensors (e.g., including or not including wrist cameras) and labels (e.g., including or not including language instructions). See~\cref{fig:datamix} and Appendix~\ref{sec:app:data_mix} for the detailed mixture. 
To create our training mixture $D$, we curate the data by first removing all Open-X datasets that contain no image streams, as well as those that do not use
delta end-effector control. We 
also remove datasets that are too repetitive, have a low image resolution, or consist of excessively niche tasks. For the remaining datasets, we roughly categorize them into ``more diverse'' and ``less diverse'' datasets based on the tasks and environments, and then double the weight of the more diverse datasets during training.
We also down-weight a few datasets with many repetitive episodes to avoid dominating the mixture. Finally, we zero-pad any missing camera channels and align the gripper action spaces between the datasets such that a gripper command of +1 means ``the gripper is open'' and 0 means ``the gripper is closed.'' While we found the resulting training mixture to work well, future work should perform a more thorough analysis of data mixture quality for pretraining general robot policies.

\begin{figure*}[t]
    \centering
    \includegraphics[width=\textwidth]{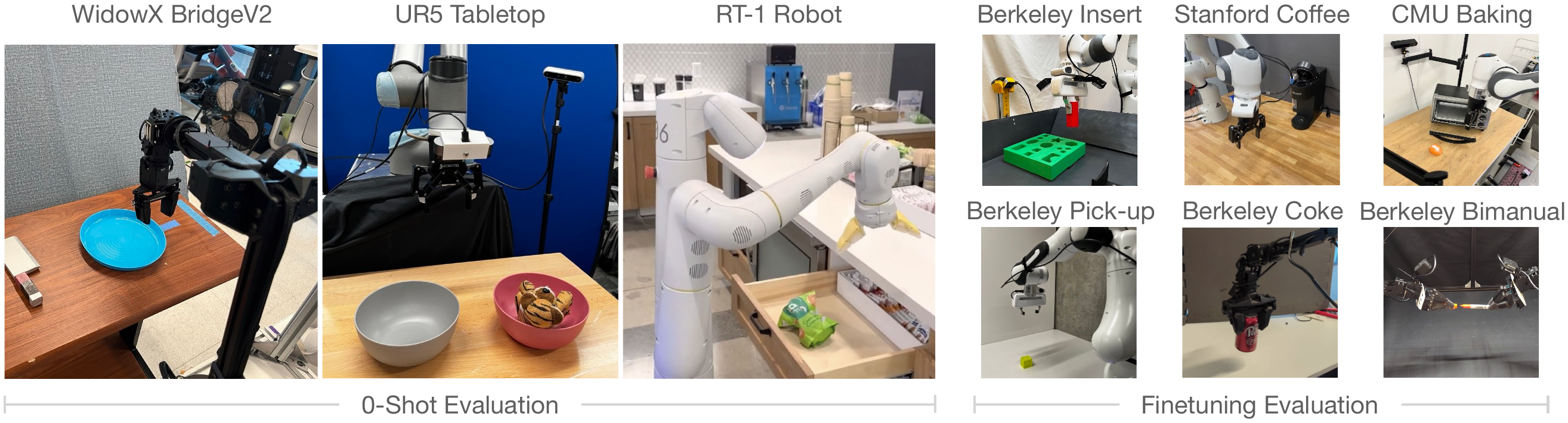}
    \caption{\textbf{Evaluation Tasks.} We evaluate \method{} on \nsetups{}~real robot setups across \ninstitutions{}~institutions. Our evaluations capture diverse object interactions (\eg ``WidowX BridgeV2''), long task horizons (\eg ``Stanford Coffee'') and precise manipulation (\eg ``Berkeley Peg Insertion''). We evaluate \method{}'s capabilities to control robots in environments from the pretraining data out-of-the-box and to efficiently finetune to new tasks and environments with small target domain datasets. We also test finetuning with new observations (force-torque inputs for ``Berkeley Peg Insertion''), action spaces (joint position control in ``Berkeley Pick-Up'' and ``Berkeley Bimanual'') and new robot embodiments (\eg ``Berkeley Bimanual'' and ``Berkeley Coke'').}
    \label{fig:eval_setups}
\end{figure*}

\subsection{Training objective}
\label{sec:train_objective}

We use a conditional diffusion decoding head to predict continuous, multi-modal action distributions~\citep{ho2020denoising,chi2023diffusionpolicy}. Importantly, only one forward pass of the transformer backbone is performed per action prediction, after which the multi-step denoising process is carried out entirely within the small diffusion head. We found this policy parameterization to outperform policies trained with MSE action heads or discretized action distributions~\citep{brohan2022rt} in both zero-shot and finetuning evaluations. To generate an action, we sample a Gaussian noise vector $x^K \sim \mathcal{N}\big(0, I\big)$ and apply $K$ steps of denoising with a learned denoising network $\epsilon_\theta(x^k, e, k)$ that is conditioned on the output $x^k$ of the previous denoising step, the step index $k$, and the output embedding $e$ of the transformer action readout: 
\begin{equation}
    x^{k-1} = \alpha (x^k - \gamma \epsilon_\theta(x^k, e, k) + \mathcal{N}\big(0, \sigma^2I\big)).
\end{equation}
The hyperparameters $\alpha$, $\gamma$, and $\sigma$ correspond to the noise schedule: we use the standard cosine schedule from~\cite{nichol2021improved}.
We train the diffusion head using the standard DDPM objective first proposed in \cite{ho2020denoising}, where we add Gaussian noise to the dataset actions and train the denoising network $\epsilon_\theta(x^k, e, k)$ to reconstruct the original action.
For a detailed explanation of diffusion policy training, see \citet{chi2023diffusionpolicy}. We list all hyperparameters in Appendix~\ref{sec:app:hyperparams}.

We use the same diffusion training objective during finetuning and update the full model, a recipe which outperformed those that freeze subsets of the pretrained parameters. In all finetuning experiments, we employ the same recipe: given a small target domain dataset with around 100 trajectories, we finetune for 50k steps using a cosine decay learning rate decay with linear warmup.

\subsection{Training Details}
\label{sec:training_details}
We trained two variants of our model: \method-Small with a transformer backbone that mirrors the size of a ViT-S, and \method-Base with a transformer backbone that mirrors the size of a ViT-B~\citep{dosovitskiy2020image}. 

We use the AdamW optimizer~\citep{loshchilov2017decoupled} with an inverse square root decay learning rate schedule~\citep{zhai2022scaling}, with weight decay of 0.1 and gradient clipping of 1.0. The ViT-B was trained for 300k~steps with a batch size of $2048$ using a TPU v4-128 pod, which took 14~hours. A finetuning run of the same model on a single NVIDIA~A5000 GPU with 24GB of VRAM takes approximately 5~hours and can be sped up with multi-GPU training. 

We train using 2 frames of observation history; in our preliminary experiments, we found significantly diminishing gains beyond the first additional frame. We use hindsight goal relabeling~\citep{andrychowicz2017hindsight}, which selects a state uniformly from the future in the trajectory to assign as the goal image, similar to prior work \citep{lynch2020language, walke2023bridgedata, shah2023vint,rosetebeas2022latent,mees2023grounding}. We apply common image data augmentations during training, and randomly zero out the language instruction or goal image per training example to enable \method{} to be conditioned on \emph{either} language instructions \emph{or} goal images. For datasets without language annotations, we always use goal image conditioning. This enables our model to learn control mostly from self-supervised visual observations and reduces the burden on language annotation, similar to prior work on multi-context imitation learning~\citep{lynch2020language,mees2022calvin,mees2022matters,mees2023grounding}. For more details on the choice of hyperparameters, see Appendix~\ref{sec:app:hyperparams}.

\subsection{Model Checkpoints \& Code}
\label{sec:code}

We open-source all resources required to train, finetune and run our model (see \website):
\begin{itemize}
    \item \textbf{Pretrained \method{} checkpoints} for \method-Small (\smallnparams{}~params) and \method-Base (\largenparams{}~params). %
    \item \textbf{Finetuning scripts} for \method{} models, in JAX.
    \item \textbf{Model pretraining pipeline} for \method{} pretraining on the Open X-Embodiment dataset, in JAX.
    \item \textbf{Standalone data loaders} for Open X-Embodiment data, compatible with JAX and PyTorch.
\end{itemize}
We provide a simple example for loading and running a pretrained \method{} model in Appendix~\ref{sec:app:code_example}.

%% file: sections/04_experiments.tex
\section{Experiments}
\label{sec:experiments}

Our experiments provide an empirical analysis of \method{}, evaluating its ability to serve as a general robotic foundation model across several axes:
\begin{enumerate}
    \item Can \method{} control multiple robot embodiments and solve language and goal tasks out of the box?
    \item Do \method{} weights serve as a good initialization for data-efficient finetuning to new tasks and robots, and does it improve over training from scratch and commonly used pretrained representations?

    \item Which design decisions in \method{} matter most for building generalist robot policies?
\end{enumerate}

\paragraph{Evaluation setups} We evaluate \method's capabilities across a representative spectrum of \nsetups{}~robot learning setups at \ninstitutions{}~institutions (see \cref{fig:eval_setups}).
We test \method's ability to control different robots out-of-the-box (``zero-shot'') for language and goal image tasks using robot setups that match the pretraining data, where all robots are controlled with delta end-effector control actions and the observation spaces are RGB images. We also evaluate \method{} for data-efficient finetuning to new environments and tasks, including with new observations (force-torque inputs in ``Berkeley Insertion''),  new action spaces (joint position control in ``Berkeley Pick-Up'') and new robot embodiments (``Berkeley Coke'' and ``Berkeley Bimanual'').
Each of the finetuning setups uses $\sim100$ in-domain demonstrations and finetunes in $<5$~hours on a NVIDIA~A5000 GPU, using the same hyperparameters across all setups (see Appendix~\ref{sec:app:hyperparams}). Our evaluation tasks test \method's ability to interact with diverse objects (\eg ``WidowX BridgeV2''), solve long-horizon tasks (\eg ``Stanford Coffee'') and perform precise manipulation (\eg ``Berkeley Insertion''). For more details on each evaluation setup, see Appendix~\ref{sec:app:robot_setups}.

\paragraph{Comparisons} We compare \method's ability to control multiple robots out-of-the-box to the best openly available generalist robot policy, \textbf{RT-1-X}~\citep{open_x_embodiment_rt_x_2023}, using the released checkpoint. Similar to \method, RT-1-X is pretrained on the Open X-Embodiment robot dataset and aims to control multiple robots zero-shot, thus providing a natural point of comparison. We also compare the zero-shot capabilities of \method{} to \textbf{RT-2-X}, a 55 billion parameter vision-language model finetuned on the Open X-Embodiment dataset to produce robot actions. The RT-1-X and RT-2-X models \citep{open_x_embodiment_rt_x_2023} are trained on a more restricted subset of 350K episodes (compared to \ntrajs{} episodes for \method).
We further compare \method's performance as a policy initialization for data efficient finetuning to two common approaches: (1)~training on the target domain demonstrations \emph{from scratch} and (2)~using pretrained visual representations. While a number of prior works have proposed other pretraining schemes for imitation finetuning~\citep{ebert2021bridge, Du2023BehaviorRF, fang2023rh20t}, to our knowledge no prior method provides a pretrained \emph{policy} that has been demonstrated to finetune successfully to \emph{new} observation and action spaces. However, pretrained visual representations such as VC-1~\citep{vc2023} have been used in this way, and therefore we use these methods as another point of comparison.

For finetuning, we found that training our large transformer architecture from scratch overfit quickly on the small datasets. Instead, we obtained better from-scratch results using a canonical policy architecture employed by many prior works: a ResNet visual encoder with FiLM~\citep{perez2018film} language conditioning, combined with a small transformer action decoder trained with a diffusion objective, similar to \citep{brohan2022rt,zhao2023learning,chi2023diffusionpolicy,lynch2023interactive}. Our instantiation of this architecture has 28M parameters (similar to RT-1 \cite{brohan2022rt}). We adopt this as our from-scratch baseline (\textbf{``ResNet+Transformer Scratch''}). We also compare to a pretrained visual representation following the procedure of \citet{majumdar2023we}. A ViT-B visual encoder is initialized to the VC-1 weights~\citep{majumdar2023we}, a state-of-the-art visual representation pretrained on 4,000~hours of ego-centric videos and ImageNet, and combined with an MLP action decoder. The full model is trained to predict expert actions using an MSE loss (\textbf{``VC-1''}).

\subsection{\method{} Controls Multiple Robots Out-of-the-Box}
\label{sec:exp_zero_shot}

\begin{table*}[th]
\centering
\footnotesize
\resizebox{\textwidth}{!}{\begin{tabular}{lccccccc}
\toprule
& Berkeley Insertion$^*$  & Stanford Coffee & CMU Baking  & Berkeley Pick-Up$^\dagger$ & Berkeley Coke & Berkeley Bimanual$^\dagger$ & Average\\

\midrule

ResNet+Transformer Scratch & 10\% & 45\% & 25\% & 0\% & 20\% & 20\% & 20\%\\

VC-1~\citep{majumdar2023we} & 5\% & 0\% & 30\% & 0\% & 10\% & 50\% & 15\% \\

\method{} (Ours) & \textbf{70\%} & \textbf{75\%} & \textbf{50\%} & \textbf{60\%} & \textbf{100\%} & \textbf{80\%} & \textbf{72\%}\\

\bottomrule
\end{tabular}}
\caption{\textbf{Finetuning Evaluation.} \method{} enables data-efficient finetuning to new domains and out-performs training from scratch as well as state-of-the-art pretrained visual representations. Each domain uses $\sim100$ target demonstrations and the same finetuning hyperparameters. In each domain, success rates are averaged over 20 trials. $*$: New observation input (force-torque proprioception). $\dagger$: New action space (joint position control).}
\label{tab:finetune_exps}
\end{table*}

\begin{figure}[h]
    \centering
    \includegraphics[width=0.5\textwidth]{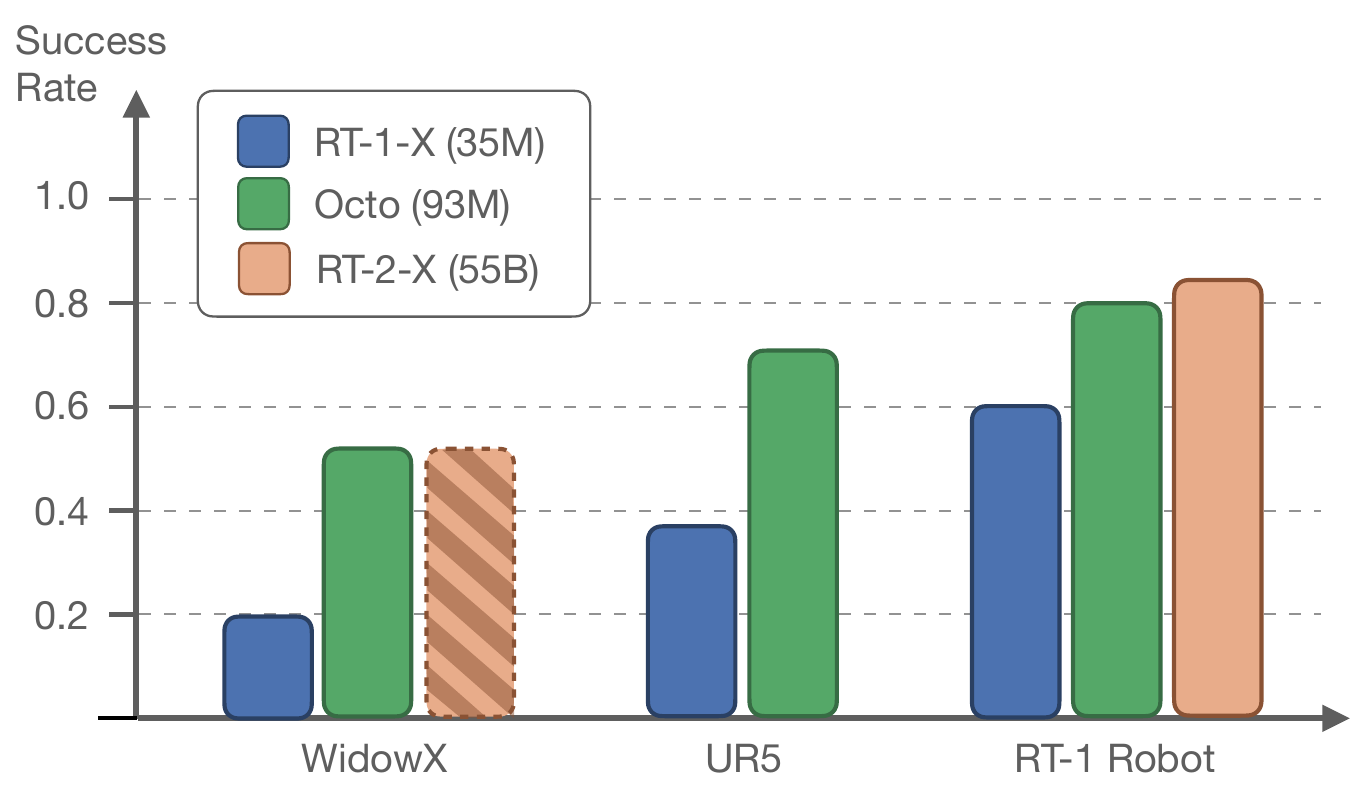}
    \caption[]{\textbf{Zero-Shot Evaluation.} Out-of-the-box, \method{} can control multiple robots in environments from the pretraining data. When using natural language to specify tasks, \method{} outperforms RT-1-X~\citep{open_x_embodiment_rt_x_2023}, the current best openly available generalist robot policy across three different robot embodiments and setups. \method{} also performs similarly to RT-2-X~\citep{zitkovich2023rt} on the tested WidowX and RT-1 Robot tasks.\textsuperscript{1} }
    \label{fig:eval_zero_shot}
\end{figure}
\footnotetext{\textsuperscript{1}For the WidowX, since RT-2-X is not openly available, we report the RT-2-X numbers from~\cite{black2023zero} (dashed bar) and use the same tasks for the Octo and RT-1-X evaluations. For the RT-1 Robot, the authors of RT-2-X kindly performed the evaluations for us.}

We compare the zero-shot manipulation capabilities of \method{}, RT-1-X, and RT-2-X in \cref{fig:eval_zero_shot}. We evaluated on several tasks selected from the pre-training dataset including picking and placing, wiping a table with a cloth, and opening and closing drawers. For each robot, we selected two language tasks from the corresponding OXE dataset and performed 10 trials per task with varying initial conditions (details in Appendix \ref{sec:app:robot_setups}). The chosen tasks are ``in-distribution'' from the pre-training data, but the evaluation requires methods to generalize to new object positions, lighting conditions, backgrounds, and distractor objects. While all methods acted reasonably across tasks in the pretraining environments, we found that on average \method{} had a 29\% higher success rate than RT-1-X (35M parameters). For the WidowX and RT-1 Robot evaluations, we also compared to RT-2-X (55 billion parameters)~\citep{zitkovich2023rt} and found that \method{} performed similarly.

Additionally, while RT-1-X and RT-2-X only support conditioning on language instructions, \method{} also supports conditioning on goal images. We evaluated our model on the WidowX tasks using goal image conditioning and found that it achieved a 25\% higher success rate than when evaluated with language conditioning. This is likely because goal images provide more information about how to achieve the task. In the BridgeV2 domain, we performed a fine-grained analysis of the zero-shot capabilities in \cref{tab:gen-type}; measuring performance on setups seen in the dataset, and for novel environments, scenes, and skills. While the \method{} model achieves high success on novel objects, zero-shot performance slightly degrades in a new scene, and high degradation for novel behaviors like flipping or precise insertion.

\subsection{\method{} Enables Data-Efficient Learning in New Domains}
\label{sec:exp_finetuning}

We report data-efficient finetuning results to new domains in \cref{tab:finetune_exps}. We find that finetuning \method{} leads to better policies than starting from scratch or with the pretrained VC-1 weights. On average across the six evaluation setups (detailed in Appendix \ref{sec:app:robot_setups}), \method{} outperforms the next best baseline by 52\%. Importantly, we use the same  recipe and hyperparameters for fine-tuning \method{} on all evaluation tasks (see \cref{sec:train_objective}),
making this a good default configuration.

The results also underline \method's ability to accommodate new observations (force-torque inputs for ``Berkeley Insertion''), action spaces (joint position control for ``Berkeley Pick-Up'') and new robot embodiments (``Berkeley Coke'' and ``Berkeley Bimanual''). This makes \method{} applicable to a wide range of single and dual arm robotic manipulation problems that go beyond a single camera input and end-effector position control.

\subsection{Design Decisions for Generalist Robot Policy Training}
\label{sec:ablations}

We have demonstrated the effectiveness of \method{} as a zero-shot multi-robot controller and as an initialization for policy finetuning. We next analyze the effects of different design decisions on the performance of the \method{} policy. Concretely, we focus on the following aspects: (1)~model architecture, (2)~training data, (3)~training objective, and (4)~model scale. Unless noted otherwise, we perform all ablations on the \method{}-Small model due to our compute budget.

\begin{table}
\centering
\resizebox{\linewidth}{!}{\begin{tabular}{ccc}
\toprule
&& Aggregate Performance\\

\midrule
&\method{}-Small (Ours) & \textbf{83\%} \\
\midrule

\parbox[t]{2mm}{\multirow{1}{*}{\rotatebox[origin=c]{90}{\textsc{data}}}}& RT-X dataset mix \vspace{0.13cm} \citep{open_x_embodiment_rt_x_2023} & 60\% \\
& Single robot dataset (Bridge Data) & 43\% \\
\midrule

\parbox[t]{2mm}{\multirow{1}{*}{\rotatebox[origin=c]{90}{\textsc{policy}}}}& Discretized Action Prediction \vspace{0.13cm} \citep{open_x_embodiment_rt_x_2023} & 18\% \\
& Continuous Action Prediction (MSE) \vspace{0.13cm} & 35\% \\
\midrule
\parbox[t]{2mm}{\multirow{1}{*}{\rotatebox[origin=c]{90}{\textsc{arch}}}}& \vspace{0.4cm}
Resnet-50 + Transformer\citep{open_x_embodiment_rt_x_2023} & 70\% \\
\bottomrule
\end{tabular}}

\caption{\textbf{Model Ablations.} We achieve best performance when using the ViT architecture, diffusion action head, and wide training data mixture. All evaluations are performed on the WidowX setup. Success rates are averaged over 40 trials across two language-conditioned tasks and two goal-conditioned tasks.}
\label{tab:ablations}
\end{table}

\paragraph{Model architecture} Prior transformer-based policy designs typically encode input images with large ResNet-style~\citep{he2016deep} encoders and fuse the resulting image features with a comparatively small transformer~\citep{brohan2022rt,open_x_embodiment_rt_x_2023,shah2023vint,chi2023diffusionpolicy,zhao2023learning,mees2022matters,shridhar2023perceiver}. Instead, we opt for a ``transformer-first'' architecture that uses very shallow CNN patch encoders and concentrates most of the parameters and FLOPS in the transformer backbone, similar to canonical vision transformer architectures~\citep{dosovitskiy2020image}. In Table \ref{tab:ablations} we show that this scalable architecture leads to substantially improved performance when training on the full Open X-Embodiment data mix. Importantly, we found ResNet-based architectures to perform better than ViTs when training on small datasets, \eg in our ``from scratch'' comparisons, underlining that large transformer policies are uniquely suited for scalable training on diverse datasets.

\paragraph{Training data} \method{} is trained on the most diverse cross-embodied robot dataset to date, a mix of \ndatasets{}~datasets that we manually curated from the Open X-Embodiment dataset~\citep{open_x_embodiment_rt_x_2023} (see \cref{sec:train_data}). We ablate the impact of this training mix by comparing to \method{} models trained on a smaller mix of 11~datasets used in training the RT-X models~\cite{open_x_embodiment_rt_x_2023} and a baseline trained only on data from the target robot domain. In Table \ref{tab:ablations} we show that the performance of \method{} increases as we increase the number of training datasets. This suggests that expanding the data mix to even more datasets may further improve policy performance. We will leave this for future work, along with a more thorough investigation of best practices for data curation.

\paragraph{Training objective} We compare \method{}'s diffusion decoding training objective (see \cref{sec:train_objective}) to common alternatives from prior work: simple MSE loss~\citep{bojarski2016end,levine2016end} and cross-entropy loss on discretized actions~\citep{brohan2022rt,zitkovich2023rt}. In Table \ref{tab:ablations} we find that \method{}'s diffusion training objective leads to substantially improved performance. This improvement is likely because the diffusion head can model multi-modal action distributions (unlike the MSE head) while retaining the precision of continuous actions (unlike the discrete head). Qualitatively, the policy acts more decisively than MSE-trained policies, and more precisely than those trained with discretized actions. 

\begin{figure}
    \centering
    \includegraphics[width=\linewidth]{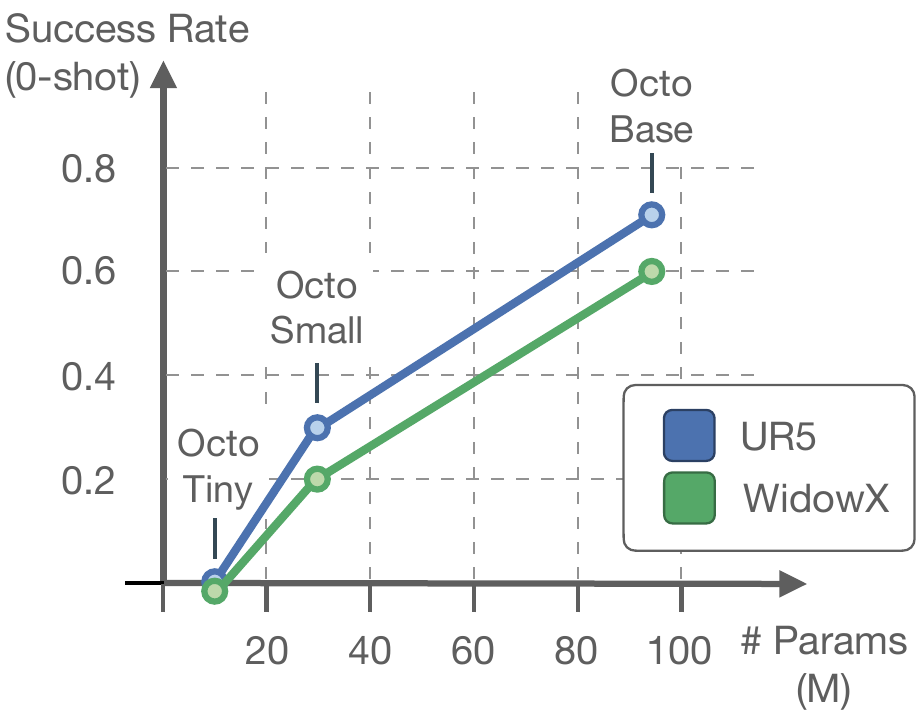}
    \caption{\textbf{Model Scaling.} The performance of \method{} improves with larger model sizes on both UR5 and WidowX tasks. Success rates are averaged over 10 trials on one language-conditioned task per robot.}
    \label{fig:scaling}
\end{figure}

\paragraph{Model scale} We compare \method{} models of three different sizes following the ladder of common vision transformer models~\citep{zhai2022scaling}: \method{}-Tiny (\tinynparams), \method{}-Small (\smallnparams{}), and \method{}-Base (\largenparams{}). In Figure \ref{fig:scaling} we show that the zero-shot performance of the policy scales with increasing model size. We find that the Base model is more robust to initial scene configuration than the Small model, and is less prone to early grasp attempts, indicating the larger model has better visual scene perception.

%% file: sections/05_conclusion.tex
\section{Discussion and Future Work}
\label{sec:discussion}

We introduced \method{}, a large transformer-based policy pretrained
on the largest robot manipulation dataset to date,  \ntrajs{} robot trajectories. We demonstrated that \method{} can solve a variety of tasks out-of-the-box and showed how \method's compositional design enables finetuning to new inputs and action spaces, making \method{} a versatile initialization for a wide range of robotic control problems. Apart from the model itself, we have released our full training and finetuning code, alongside tools that make it easier to train on large robot datasets.

While \method{} achieves strong performance in both zero-shot and finetuning evaluations, we find that the current model still has several short-comings, which we attribute in large parts to characteristics of the training data.
First, we found that the current \method{} model struggles with adequately processing wrist camera information. Often finetuning results were stronger when using only a third person camera instead of combining third person and wrist camera. Additionally, we notice a large difference between language-conditioned policy performance and goal-conditioned policy performance. In both cases, a lack of the respective modalities in the training data is the likely reason: only 27\% of the data contains wrist camera information and only 56\% of the pretraining data contains language annotations.

Expanding the data used to train \method{} is a natural avenue of improvement. Since the Open X-Embodiment dataset is comprised of optimal robot demonstrations, the current model trains via imitation; future work may consider learning from sub-optimal or online interaction data that require alternative objectives. Further, while we trained and evaluated \method{} exclusively on single and dual-arm manipulators; expanding to a wider set of robots that perform navigation or mobile manipulation would be an direction of high opportunity. 

While \method{} represents a step towards building generalist robot policies that work out-of-the-box on diverse robot setups, there remains work to improve the model, including better language conditioning, improved support for wrist cameras, and incorporating data beyond optimal demonstrations. We hope that \method{} offers a simple launchpad for researchers and practitioners to access larger robotic datasets and leverage pretrained robotics models for efficient learning of new tasks and broad generalization.

%% file: sections/06_appendix.tex
\begin{appendices}

\section{Contributions}\label{sec:contributions}

    \noindent\textbf{Dibya Ghosh:} led model development, proposed and implemented large parts of the final model design, babysitted the training runs and touched all parts of the codebase, helped with model evaluations and tech report writing.

    \noindent\textbf{Homer Walke:} led model evaluations, designed the main Bridge evaluation benchmark, contributed to the initial model implementation and ran many of the evals for this tech report.

    \noindent\textbf{Karl Pertsch:} managed the overall project, led Open-X data integration and curation, led writing of this tech report, contributed to model development and implementation and ran model evaluations for the tech report.

    \noindent\textbf{Kevin Black:} led data loading and training infrastructure, managed TPU pod training, created the project website, contributed to model development and implementation, and helped with robot evaluations and tech report writing.

    \noindent\textbf{Oier Mees:} ran countless ablations for model development, contributed to model implementation, helped with evals and writing of the tech report.

    \noindent\textbf{Sudeep Dasari:} contributed the model evaluations at CMU, experimented with pretrained encoders.

    \noindent\textbf{Joey Hejna:} contributed the model evaluations at Stanford.

    \noindent\textbf{Tobias Kreiman:} contributed model evaluations in simulated environments.

    \noindent\textbf{Ria Doshi:} contributed to explorations into cross-morphology training.
    
    \noindent\textbf{Charles Xu:} contributed the model evaluations for Berkeley Peg Insert.

    \noindent\textbf{Jianlan Luo:} contributed the model evaluations for Berkeley Peg Insert.

    \noindent\textbf{You Liang Tan:} helped diagnose and resolve bottlenecks in data loading.

    \noindent\textbf{Lawrence Yunliang Chen:} helped set up and run UR5 finetuning experiments at Berkeley AutoLab.

    \noindent\textbf{Pannag Sanketi, Quan Vuong, Ted Xiao:} contributed model evaluations on the Google Robot.
    
    \noindent\textbf{Dorsa Sadigh, Chelsea Finn, Sergey Levine:} provided guidance throughout the project and feedback on the writing of this tech report.

\section{\method{} Code Example}
\label{sec:app:code_example}

Loading a pretrained \method{} model and performing inference requires little code:

\begin{lstlisting}[language=Python, caption=Example Python code to perform inference with a pretrained Octo model., basicstyle=\tiny\ttfamily]
import jax
from octo.model.octo_model import OctoModel

model = OctoModel.load_pretrained("hf://rail-berkeley/octo-base")
print(model.get_pretty_spec()) # Print out the input-output spec
observation = {"image_primary": img}   
task = model.create_tasks(texts=["pick up the fork"])
action = model.sample_actions(
    observation, task, rng=jax.random.PRNGKey(0))

\end{lstlisting}

\section{Data mixture}
\label{sec:app:data_mix}

We list the detailed training mixture used for training the \method{} models in Table~\ref{tab:data_mix}. The sampling weights are mostly determined by the relative size of the datasets with a few manual adjustments (see Section \ref{sec:train_data}). 

\begin{table}[th]
    \centering
       \begin{tabular}{lr}
        \toprule
        \multicolumn{2}{c}{\textbf{\method{} Pretraining Dataset Mixture}}\\
        \midrule
        Fractal~\citep{brohan2022rt} & 17.0\% \\
        Kuka~\citep{kalashnikov2018qt} & 17.0\% \\
        Bridge\citep{ebert2021bridge, walke2023bridgedata} & 17.0\% \\
        BC-Z~\citep{jang2022bc} & 9.1\% \\
        Stanford Hydra Dataset~\citep{belkhale2023hydra}  & 6.0\% \\
        Language Table~\citep{lynch2023interactive} & 5.9\% \\
        Taco Play~\citep{rosetebeas2022latent,mees2023grounding} & 3.6\% \\
        Furniture Bench Dataset~\citep{heo2023furniturebench}  & 3.3\% \\
        UTAustin Mutex~\citep{shah2023mutex} & 3.0\% \\
        Austin Sailor Dataset~\citep{nasiriany2022sailor}  & 2.9\% \\
        Roboturk~\citep{DBLP:journals/corr/abs-1811-02790} & 2.8\% \\
        Toto~\citep{zhou2023train} & 2.4\% \\
        Austin Sirius Dataset~\citep{liu2022robot}  & 2.3\% \\
        Berkeley Autolab UR5~\citep{BerkeleyUR5Website} & 1.5\% \\
        IAMLab CMU Pickup Insert~\citep{saxena2023multiresolution}  & 1.2\% \\
        Viola~\citep{zhu2023viola} & 1.2\% \\
        Berkeley Fanuc Manipulation~\citep{fanuc_manipulation2023} & 1.0\% \\
        NYU Franka Play Dataset~\citep{cui2022play}  & 0.9\% \\
        UCSD Kitchen Dataset~\citep{ucsd_kitchens}  & <0.1\% \\
        Jaco Play~\citep{dass2023jacoplay} & 0.6\% \\
        Berkeley Cable Routing~\citep{luo2023multistage} & 0.3\% \\
        Austin Buds Dataset~\citep{zhu2022bottom}  & 0.3\% \\
        CMU Stretch~\citep{mendonca2023structured} & 0.2\% \\
        NYU Door Opening~\citep{pari2021surprising}  & 0.1\% \\
        DLR EDAN Shared Control~\citep{quere_shared_2020}  & 0.1\% \\   
                \bottomrule
        \end{tabular}%
        \caption{\method{} pretraining data mixture using datasets from the Open X-Embodiment dataset~\citep{open_x_embodiment_rt_x_2023}.}
        \label{tab:data_mix}
\end{table}

\section{Training Hyperparameters}
\label{sec:app:hyperparams}

We mostly follow documented practices for training vision transformers~\citep{zhai2022scaling}. We use the AdamW optimizer~\citep{loshchilov2017decoupled} with an inverse square root decay learning rate schedule~\citep{zhai2022scaling} and linear learning rate warm-up. We list hyperparamaters used during training in Table~\ref{tab:hyperparams} and the model parameters for the different sizes in Table~\ref{tab:models}. We apply standard image augmentations during training. Concretely, for the 3rd person camera we apply stochastic crops followed be a resize to $256 \times 256$, followed by color jitter. Finally, we normalize the input image to have pixels with float values between -$1.0$ and $1.0$. For the wrist camera, we apply the same procedure except without the random crop and resizing to $128 \times 128$ instead.
\begin{table}[th]
\centering
\begin{tabular}{cc}
\toprule
Hyperparameter
     &  Value \\
     \midrule
  Learning Rate   & 3e-4\\
  Warmup Steps & 2000\\
  LR Scheduler & reciprocal
square-root\\
  Weight Decay & 0.1\\
  Gradient Clip Threshold & 1\\
  Batch Size & 2048\\
  \bottomrule
\end{tabular}
\caption{Hyperparameters used during training.}
\label{tab:hyperparams}
\end{table}

The images are passed through a shallow convolution stack, then split into a sequence of flattened patches~\citep{dosovitskiy2020image} of size $16 \times 16$. This results in 256 tokens for the 3rd person camera images and 64 tokens for the wrist camera images. For datasets containing language annotations, we use a pretrained  \texttt{t5-base} (111M) transformer model \citep{2020t5} that produces a sequence of 16 language embedding tokens.
\begin{table}[th]
\centering
\small
\setlength{\tabcolsep}{3pt}
\begin{tabular}{l c c c c c}
\toprule
Model            & Layers & Hidden size $D$ & MLP size &  Heads  & Params \\
\midrule 
\method-Small   &   12   &        384      &   1536   &   6    &    27M \\
\method-Base  &   12   &       768      &   3072   &   12    &  93M  \\

\bottomrule
\end{tabular}
\caption{Architecture details of \method{} model variants.}
\label{tab:models}
\end{table}

The diffusion action head consists of a 3-layer MLP with a hidden dimension of 256, residual connections, and layer normalization. We use the standard DDPM objective as introduced by~\citep{ho2020denoising} with a cosine noise schedule~\citep{nichol2021improved} and 20 diffusion steps.

\section{Things that Worked and Did Not Work (Yet)}
\label{sec:app:things_that_did_not_work}

\paragraph{Things we found improved performance}
\begin{itemize}
    \item \textbf{Adding history during pretraining}: Models with one frame of history as context performed better in zero-shot evals than models pretrained without history. We did not observe benefits of increasing the history length further on the few tasks we evaluated on, though other tasks may benefit.
    \item \textbf{Using action chunking}: We found it helpful to use ``action chunking''~\citep{zhao2023learning}, \ie to predict multiple actions into the future, for getting more coherent policy movements. In our evaluations, we did not find temporal ensembling of future actions to provide additional benefits beyond receding horizon control.
    \item \textbf{Decreasing patch size} Tokenizing images into patches of size $16 \times 16$  led to improved performance over patches of size $32 \times 32$, particularly for grasping and other fine-grained tasks. This does add compute complexity (the number of tokens is $4 \times$), so understanding how to balance compute costs and resolution remains a problem of interest.
    \item \textbf{Increasing shuffle buffer size}: Loading data from 25 datasets in parallel is a challenge. Specifically, we found that achieving good shuffling of frames during training was crucial --- zero-shot performance with a small shuffle buffer (20k) and trajectory-level interleaving suffered significantly.
    We solved this issue by shuffling and interleaving frames from different trajectories \textit{before} decoding the images, allowing us to fit a much larger shuffle buffer (up to 500k). We also subsample at most 100 randomly chosen steps from each training trajectory during data loading to avoid ``over-crowding'' the shuffle buffer with single, very long episodes.
\end{itemize}

\paragraph{Things that did not work (yet)}
\begin{itemize}
    \item \textbf{MSE Action Heads}: Replacing our diffusion decoding head with a simple L2 loss led to ``hedging'' policies that move very slowly and \eg fail to rotate the gripper in WidowX evaluations.
    \item \textbf{Discrete Action Heads}: Discretizing actions into 256 bins per dimension and training with cross-entropy loss like in~\citet{brohan2022rt} led to more ``decisive'' policies, yet they lacked precision and often missed the grasp.
    \item \textbf{ResNet Encoders}: did not scale as well to larger datasets in our evaluations (see Table \ref{tab:ablations}), though they did outperform our ViT architecture when training from scratch on a small dataset (around 100 demonstrations).
    \item \textbf{Pretrained Encoders}: ImageNet pretrained ResNet encoders did not provide benefit on zero-shot evals, though may be confounded with ResNet architectures underperforming as mentioned above.
    \item \textbf{Relative Gripper Action Representation}: When aligning the gripper action representations of the different datasets, we tried (A) absolute gripper actions, \ie actions are +1 when the gripper is open and 0 if it is closed, and (B) relative gripper actions, \ie gripper action is +1/0 only in the timestep when the gripper opens/closes and 0.5 otherwise. We found that the latter tends to open/close the gripper less often since most of the training data represents ``do not change gripper'' actions, leading to a slightly higher grasp success rate. At the same time, the relative representation led to less retrying behavior after a grasp failed, which was ultimately worse. Thus, we chose the absolute gripper action representation.
    \item \textbf{Adding Proprioceptive Inputs}: Policies trained with propioceptive observations seemed generally worse, potentially due to a strong correlation between states and future actions. We hypothesize this might be due to a causal confusion between the proprioceptive information and
the target actions~\citep{de2019causal}.
    \item \textbf{Finetuning Language Model}: In order to improve the visuo-lingual grounding of \method{} we experimented with: i)~varying sizes of the T5 encoder~\citep{2020t5}: \texttt{small}~(30M), \texttt{base}~(111M), and \texttt{large}~(386M) as well as ii)~finetuning the last two layers of the encoder. Using the frozen \texttt{base} model resulted in the best language-conditioned policies. We did not find improvements when using larger encoders or finetuning the encoder. We hypothesize this might be due to the lack of rich, diverse, free-form language annotations in most of the datasets.
    
\end{itemize}

\section{Experimental Setups}
\label{sec:app:robot_setups}

\begin{figure*}[h]
    \centering
    \includegraphics[width=\textwidth]{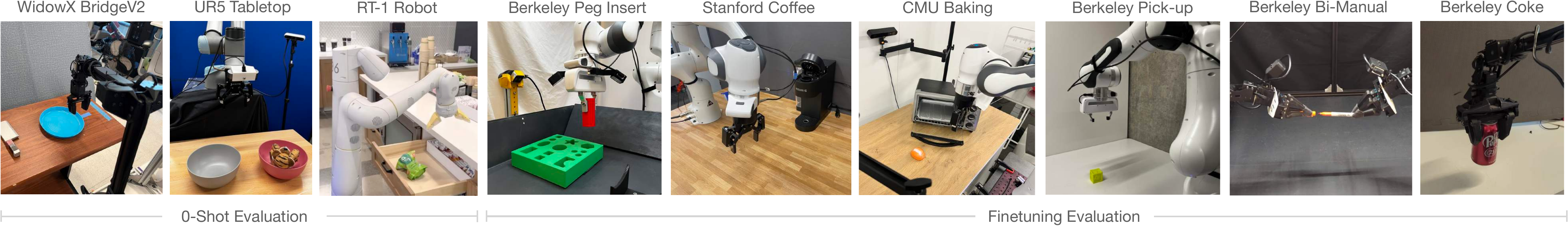}
    \caption{\textbf{Evaluation Tasks.} Replicated from the main text for convenience. We evaluate \method{} on \nsetups{}~real robot setups across \ninstitutions{}~institutions in zero-shot and finetuning scenarios.}
    \label{fig:eval_setups_2}
\end{figure*}

\subsection{Zero-Shot Evaluations}

\paragraph{WidowX BridgeV2} Uses the setup of \citet{walke2023bridgedata}, in which a Trossen WidowX 250 6-DOF robot performs diverse manipulation tasks. The observation consists of a single third person camera stream and the action space is end-effector position deltas. We evaluated two language-conditioned tasks in which a the robot needs to ``place the carrot on plate'', and ``put the eggplant in the pot.'' While these tasks are in-distribution, the policy must still generalize to novel object positions. We performed 10 trials per task and varied objects positions between trials.  

\paragraph{UR5} Uses the setup of \citet{BerkeleyUR5Website}, in which a UR5 robot arm performs multiple table top manipulation tasks. The observation consists of a single third person camera stream and the action space is end-effector position deltas. We evaluated twp language-conditioned tasks: picking a toy tiger from a bowl and placing it into a different bowl as well as wiping a table with a cloth. While these tasks are in-distribution, the policy must still generalize to novel object positions, distractor objects, and lighting. Since the training data was collected months ago and the robot setup was taken down and re-assembled, the policy must also generalize to other miscellaneous changes in the environment like a slightly different camera view and background. We performed 10 trials per task and varied objects positions between trials.

\paragraph{RT-1 Robot} Uses the setup of \citet{brohan2022rt}, in which a proprietary robot performs multiple table top and furniture manipulation tasks. The observation consists of a single third person camera stream and the action space is end-effector position deltas. We evaluated on the task of picking up a 7up can, apple, blue chip bag, or brown chip bag, as well as the task of opening or closing drawers on a cabinet. While these tasks are in-distribution, the policy must still generalize to novel object positions. Additionally, since the robot setup was moved between buildings, the policy must also generalize to miscellaneous changes in the environment like a slightly different camera view and background. We performed 10 trials per task and varied objects positions between trials.

\subsection{Model Ablations}
All of our model ablations were evaluated on the WidowX setup. We present a more detailed breakdown of the success rates per task in Table \ref{tab:ablations-breakdown}. We evaluated on two language-conditioned tasks (put carrot on plate and put eggplant in pot) and two goal-conditioned tasks (put bread on plate and put spoon on glove). The goal-conditioned tasks contain objects not seen in the Bridge dataset. Additionally, we analyze the generalization capabilities of Octo across several different axes, such as novel objects, novel environments and novel skills in Table \ref{tab:gen-type}.

\begin{table*}
\centering
\resizebox{\textwidth}{!}{\begin{tabular}{ccccccc}
\toprule
&& \multicolumn{2}{c}{Language-conditioned} & \multicolumn{2}{c}{Goal-conditioned} & \\
\cmidrule(lr){3-4}
\cmidrule(lr){5-6}
&& Put carrot on plate & Put eggplant in pot & Put bread on plate & Put spoon on glove & Average \\
\midrule

&\method{}-small (Ours) & 80\% & 90\% & 70\% & 90\% & \textbf{83\%} \\
\midrule

\parbox[t]{2mm}{\multirow{1}{*}{\rotatebox[origin=c]{90}{\textsc{data}}}} & RT-X dataset mix \citep{open_x_embodiment_rt_x_2023} & 80\% & 80\% & 40\% & 40\% & 60\% \\
& Single robot dataset (Bridge Data) & 20\% & 70\% & 60\% & 20\% & 43\% \\
\midrule

\parbox[t]{2mm}{\multirow{1}{*}{\rotatebox[origin=c]{90}{\textsc{policy}}}}& Discretized Action Prediction \vspace{0.13cm} \citep{open_x_embodiment_rt_x_2023} & 0\% & 20\% & 10\% & 40\% & 18\% \\
& Continuous Action Prediction (MSE) \vspace{0.13cm} & 70\% & 30\% & 0\% & 40\% & 35\% \\
\midrule

\parbox[t]{2mm}{\multirow{1}{*}{\rotatebox[origin=c]{90}{\textsc{arch}}}}& \vspace{0.4cm} Resnet-50 + Transformer \citep{open_x_embodiment_rt_x_2023} & 80\% & 60\% & 100\% & 40\% & 70\% \\
\bottomrule
\end{tabular}}

\caption{\textbf{Model Ablations.} We achieve best performance when using the ViT architecture, diffusion action head, and wide training data mixture. All evaluations are performed on the WidowX setup. Success rates are averaged over 40 trials across two language-conditioned tasks and two goal-conditioned tasks.}
\label{tab:ablations-breakdown}
\end{table*}

\begin{table*}
\centering
{
\begin{tabular}{cccc}
\toprule

Generalization Type & Task & Success Rate & Average  \\
\midrule

\multirow{2}{*}{In-distribution} & Put carrot on plate & 80\% & \multirow{2}{*}{85\%}\\
& Put eggplant in pot & 90\% \\
\midrule

\multirow{2}{*}{Novel objects} & Put bread on plate & 70\% & \multirow{2}{*}{80\%} \\
& Put spoon on glove & 90\% \\
\midrule

\multirow{2}{*}{Novel environment} & Put mushroom in pot & 20\% & \multirow{2}{*}{40\%} \\
& Put spoon on cloth & 60\% \\
\midrule

\multirow{2}{*}{Novel skill} & Flip cup on its side & 10\% & \multirow{2}{*}{5\%} \\
& Put block in slot & 0\%\\
\bottomrule

\end{tabular}}

\caption{\textbf{Zero-shot Generalization Analysis.} Using the WidowX setup, we analyze the generalization capabilities of \method-small along several different axes. \method{} performs the best on in-distribution tasks and generalizes well to novel objects and environments. However, \method{} struggles to generalize to skills not seen in the WidowX embodiment. Success rates are averaged over 20 trials across two tasks.}
\label{tab:gen-type}
\end{table*}

\subsection{Finetuning Evaluations}
\paragraph{CMU Baking} The robot must pick up the toy bread object, place it in the toaster, and shut the toaster. This task requires generalization across initial positions (of both the toaster and object) and the shape of the target toy bread object. We use an end-effector delta action space (Cartesian position + rotation delta). Observations come from the 3rd-person front-facing Zed camera. Actions are predicted at 15 Hz, and executed on the robot using the \href{https://github.com/AlexanderKhazatsky/R2D2}{R2D2 Franka controller}. The finetuning dataset consists of 120 demos collected via expert VR tele-operation, and every policy was evaluated using 20 trials (4 novel test objects with 5 positions each). 

\paragraph{Stanford Coffee} The robot is tasked with picking up one of four different Keurig Coffee Pods and placing it inside of a Keurig machine. This task requires both generalization across initial positions and colors of the coffee pod, as well as precise placement in the Keurig machine. We use an end effector delta action space with an open source controller running at 10 Hz based on Polymetis~\citep{Polymetis2021} (found \href{https://github.com/jhejna/robot-lightning}{here}). We use only a single 3rd-person wrist observation. Our training dataset contained 118 expert demonstrations from varied coffee pods and positions collected via VR tele-operation. We evaluated policies for 20 episodes, five episodes for each of four different color coffee pods.

\paragraph{Berkeley Peg Insertion} The task is to insert a pre-grasped 3D-printed peg into a matching slot on a 3D-printed board inside the bin. The matching tolerance between the peg and the hole is 1.5mm; which makes it a contact-rich precise part-mating task. 
The robot must learn an appropriate policy to ``search" for the matching opening through contact, which necessitates the use of force/torque measurements. The observation space of the policy consists of a single side-view camera image, the end-effector twist, and the end-effector force/torque reading. The policy sends action commands as the robot's end-effector twists at 5 HZ, tracked at 1000 HZ by a low-level impedance controller. Our finetuning dataset is composed of 100 human demonstrations from the FMB dataset~\citep{luo2023fmb}. We evaluated trained policies for 20 trials with randomized board positions.

\paragraph{Berkeley Pick Up} We use the setup of \citet{radosavovic2023robot}. The robot needs to pick up a block from a table top surface after being trained on a dataset of 100 pickups of various objects. The robot uses joint position control with an underlying Polymetis control stack~\citep{Polymetis2021} (\href{https://github.com/jhejna/robot-lightning}{here}). It is conditioned on a wrist camera input image, as well as the proprioceptive readings of the robot. We evaluted on the task of picking up the yellow cube and used 20 trials.

\paragraph{Berkeley Coke} A Trossen Robotics ViperX robot is tasked with picking up a coke can from a table. The ViperX is completely unseen in the training mixture, thus this task tests the model's capabilities to generalize to a new robot embodiment. The task also requires generalization across different positions of the coke can with the ViperX robot using end-effector delta control at 5Hz.
The policy uses image observations from a third person and a wrist camera. Our training dataset contained 115 expert demonstrations collected via VR tele-operation from varied positions.

\paragraph{Berkeley Bimanual} The task requires an ALOHA bimanual robot~\cite{zhao2023learning}, consisting of two ViperX robot arms, to pick up a sharpie marker with the right hand from the workbench and remove its cap with the left hand. Since \method{} was only pre-trained with single-arm robot data, we re-initialize a new action head that maps to the 14-dimensional action space of ALOHA (2x 6 joint positions + 2x gripper position). A successful policy needs to use image observations from the left and right wrist cameras to predict precise manipulation behaviors. Best performance for both our methods and baselines was achieved by predicting an action chunk of 64 during training, and then during test-time executing 12 actions with receding horizon control before re-planning. We evaluated the uncapping task using 10 trials, where the pen was placed in a different location for each trial.

\end{appendices}